\def\year{2022}\relax
\documentclass[letterpaper]{article} 
\usepackage{aaai22}  
\usepackage{times}  
\usepackage{helvet}  
\usepackage{courier}  
\usepackage[hyphens]{url}  
\usepackage{graphicx} 
\usepackage{natbib}  
\usepackage{caption} 
\DeclareCaptionStyle{ruled}{labelfont=normalfont,labelsep=colon,strut=off} 
\usepackage{algorithm}
\usepackage{algorithmic}
\usepackage{graphicx}
\usepackage{subfigure}
\usepackage{multirow}
\usepackage{tabularx}
\usepackage{amsmath}
\usepackage{amssymb}
\usepackage{bbding}
\usepackage{url}
\usepackage{multicol}
\usepackage{booktabs}
\usepackage[table,xcdraw,dvipsnames]{xcolor}
\usepackage{caption}
\usepackage{newfloat}
\usepackage{listings}
\floatstyle{ruled}
\newfloat{listing}{tb}{lst}{}
\floatname{listing}{Listing}
\title{Interventional Multi-Instance Learning \\with Deconfounded Instance-Level Prediction}
\author {
    Tiancheng Lin\textsuperscript{\rm 1,2},
    Hongteng Xu\textsuperscript{\rm 3,4,5},
    Canqian Yang\textsuperscript{\rm 1,2}
    and
    Yi Xu\textsuperscript{\rm 1,2*}
}
\affiliations {
    
    \textsuperscript{\rm 1} Shanghai Key Lab of Digital Media Processing and Transmission, Shanghai Jiao Tong University \\
    \textsuperscript{\rm 2} MoE Key Lab of Artificial Intelligence, AI Institute, Shanghai Jiao Tong University \\
    \textsuperscript{\rm 3} Gaoling School of Artificial Intelligence, Renmin University of China \\
	\textsuperscript{\rm 4} Beijing Key Laboratory of Big Data Management and Analysis Methods\quad 
	\textsuperscript{\rm 5} JD Explore Academy \\
	\{ltc19940819, charles.young, xuyi\}@sjtu.edu.cn, hongtengxu@ruc.edu.cn
}

\begin{document}

\maketitle

\begin{abstract}
When applying multi-instance learning (MIL) to make predictions for bags of instances, the prediction accuracy of an instance often depends on not only the instance itself but also its context in the corresponding bag. 
From the viewpoint of causal inference, such bag contextual prior works as a confounder and may result in model robustness and interpretability issues.
Focusing on this problem, we propose a novel interventional multi-instance learning (IMIL) framework to achieve deconfounded instance-level prediction.
Unlike traditional likelihood-based strategies, we design an Expectation-Maximization (EM) algorithm based on causal intervention, providing a robust instance selection in the training phase and suppressing the bias caused by the bag contextual prior. 
Experiments on pathological image analysis demonstrate that our IMIL method substantially reduces false positives and outperforms state-of-the-art MIL methods. 
\end{abstract}

\section{Introduction}
In many real-world scenarios, fine-grained labels of data, $e.g.$, pixel-wise annotations of high-resolution images, are often unavailable due to the limitations of human resources, time, and budgets.
To mitigate the requirement for high-quality labels, multi-instance learning (MIL) treats multiple instances as a bag and learns an instance-level predictive model from a set of labeled bags~\cite{SMI}. 
Such a paradigm has been widely used in many applications, $e.g.$, image classification~\cite{imgcls}, object detection~\cite{objdet}, semantic segmentation~\cite{xu2019camel}, $etc$. 
Among them, whole slide pathological image (WSI) classification is a representative example. 
Each WSI is a bag with a pathological label, and the patches of the WSI are unlabeled instances in the bag. 
The MIL framework learns an instance-level classifier to indicate the patches corresponding to the lesions. 

\begin{figure}[t]
     \centering
     \subfigure[]{
     \includegraphics[height=2.5cm]{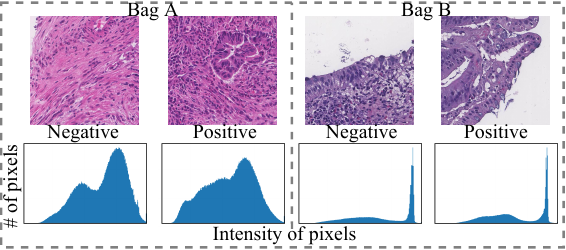}\label{fig: example and histogram}
     }
     \subfigure[]{
     \includegraphics[height=2.5cm]{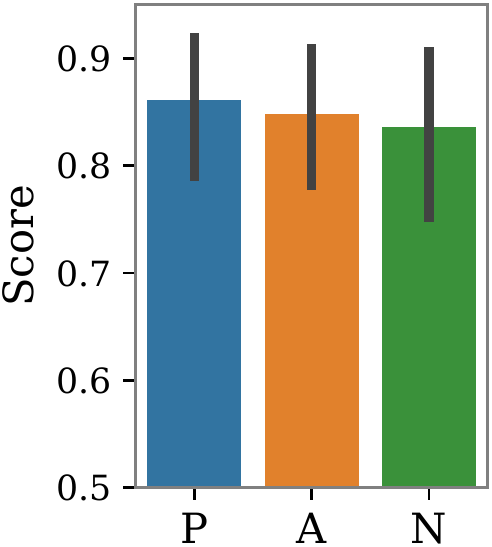}\label{fig: bar plot}
     }
    \caption{Qualitative and quantitative evidence for bag contextual prior. 
    (a) The patches in two bags and their histograms of pixel intensities.
    (b) Average scores of positive/all/negative (P/A/N) instances over positive bags and their confidence intervals.}
        \label{fig:statistic}
\end{figure}

Although many MIL methods have been proposed and achieved encouraging performance in extensive applications~\cite{hou2016cvpr,RCE,chikontwe2020topk}, they often suffer from an issue called ``bag contextual prior''. 
In particular, the bag contextual prior is a kind of instance-shared information corresponding to bags but irrelevant to their instances, which may be inherited by models like deep neural networks (DNNs) and lead to questionable instance-level prediction.
Figure~\ref{fig: example and histogram} illustrates the bag contextual prior in WSI classification. 
In each bag, its patches (instances) with different labels (positive/negative) often have similar attributes on color and texture. 
For different bags, on the contrary, their patches with the same labels can be very different in vision. 
As a result, the ``similarity'' within each bag and the ``difference'' across different bags, which harm instance-level prediction, could be wrongly exploited by MIL models. 
As shown in Figure~\ref{fig: bar plot}, the models predict similar scores for the instances in the same bag. 
From the viewpoint of causal inference, the above bag contextual prior is a confounder that causes the spurious correlation between instances and labels, making the prediction depend on both the key instance and its useless context. 
Therefore, a robust and interpretable MIL model should build an efficient mechanism to suppress the bias caused by the bag contextual prior, predicting the classification scores via revealing the actual causality between instances and labels.

\begin{figure*}[t]
    \centering
    \includegraphics[width=15.7cm]{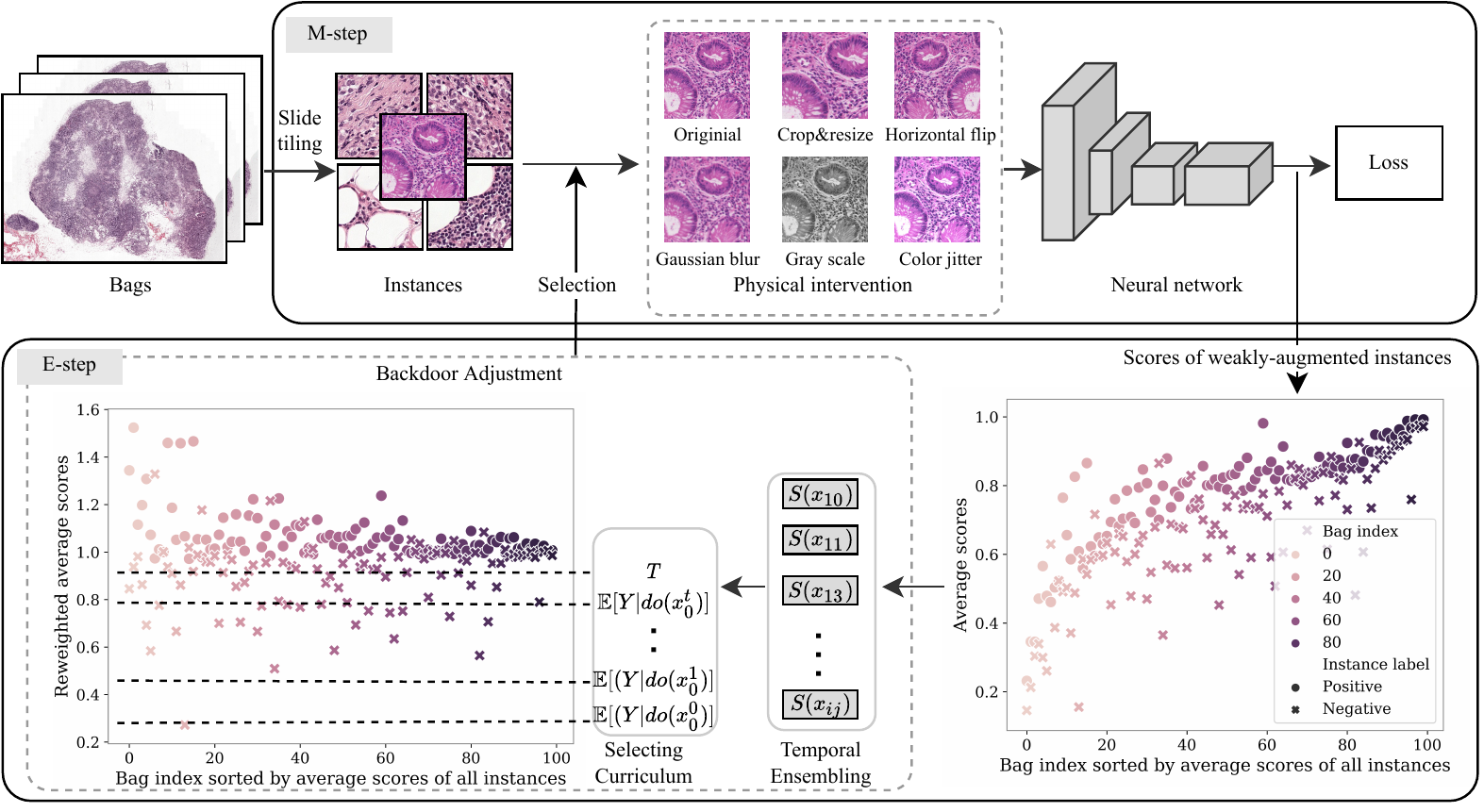}
    \caption{An illustration of the proposed IMIL framework in WSI classification.
    Physical intervention and backdoor adjustment are adopted for the deconfounded training in the M-step and the instance selection in the E-step, respectively. 
    Here, the instance-level labels are only used for illustration but never applied in the training phase.
    }
    \label{workflow}
\end{figure*}

To achieve deconfounded instance-level prediction, we propose a novel interventional multi-instance learning (IMIL), in which a structural causal model (SCM)~\cite{pearl2016causal} analyzes the causalities among the bag contextual prior, instances, and labels. 
As depicted in Figure~\ref{workflow}, our IMIL is an Expectation-Maximization (EM) algorithmic framework with two novel strategies for confounder bias removal and robust instance selection. 
In the training phase, we initially assign the instances with the label of the bag they belong to, then we alternate the following E-step and M-step until convergence.
In the M-step, we apply physical interventions to remove the confounder bias, where various data augmentations are adopted. 
In the E-step, we approximate the total effect of our model, where we first reweight the scores of instances to get the de-biased prediction and then select instances via both direct causal effect and indirect mediator effect.
Note that, different from the existing instance selection criteria~\cite{hou2016cvpr, RCE, wang2019weakly}, our model approximates the causal effect without external information.

We compare our IMIL framework with state-of-the-art MIL methods through the lens of causality and analyze their connections and differences in detail. 
The effectiveness of our IMIL is verified over two public WSI datasets, $i.e.$, DigestPath~\cite{digestpath} and Camelyon16~\cite{CAMELYON16}. 
Experimental results show that our IMIL achieves superior performance in the WSI classification tasks. 
Significantly, the proposed physical intervention is compatible with all compared MIL methods, bringing consistent performance boosting. 
Furthermore, we demonstrate the potentials of IMIL to multi-class multi-label MIL problems on the Pascal VOC dataset~\cite{everingham2015pascal}.

\section{Related Work}
\subsubsection{Bag-level MIL} 
Bag-level MIL either implicitly utilizes bag-to-bag distance/similarity or explicitly trains a bag classifier~\cite{wang2018revisiting}, whose novel distance metrics and aggregation operators are designed based on various neural network architectures~\cite{nazeri2018two,wang2019BSN,zhao2020predicting}, pooling strategies~\cite{yan2018dynamic}, and attention mechanisms~\cite{AMIL,shi2020loss}. 
For large-scale MIL scenarios like gigapixel image analysis, we often have to implement the bag-level MIL models by a two-stage strategy~\cite{campanella2019RNN,tellez2019NIC,li2019twostage,cvpr2020twostage,yao2020whole,cvper2021twostage}, training an instance-level feature extractor, and then aggregating instance features as bag-level representations. 

\subsubsection{Instance-level MIL}
Instance-level MIL is a natural solution to gigapixel image analysis, where a classifier is trained to produce a \textit{score} for each instance, and the instance scores are aggregated to produce a bag score. 
The representative method is SimpleMIL, which directly propagates the bag label to its instances ~\cite{ray2005simplemil,SimpleMIL}. 
To suppress the noise caused by the instance-level supervision, the work in~\cite{wang2019weakly} directly introduces extra cleaner annotations for partial instances by imposing larger weights on them.
Alternatively, various modifications have been introduced to SimpleMIL~\cite{hou2016cvpr,RCE,chikontwe2020topk}, aiming at using discriminative instances for training. 
As shown in Figure~\ref{workflow}, these approaches are essentially in the EM framework, training model in M-step and selecting instances in E-step.

\subsubsection{Causal Inference in Computer Vision}
Causal inference has been introduced to a growing number of computer vision tasks, including class-incremental learning~\cite{CIL}, long-tailed classification~\cite{tang2020long}, unsupervised representation learning~\cite{wang2020visual,mitrovic2020representation}, few-shot learning~\cite{yue2020interventional}, weakly supervised semantic segmentation~\cite{zhang2020causal} and so on. 
It not only offers an interpretation framework to analyze visual problems, but also empowers the models by removing the spurious correlation~\cite{qi2020two}, leveraging path-effect analysis~\cite{niu2021counterfactual}, and generating counterfactual examples~\cite{yue2021counterfactual}. 
In MIL problems, StableMIL~\cite{Zhang2019StableML} takes ``adding an instance to a bag'' as a treatment and observes the  bag labels under different treatments, which focuses on bag-level prediction, while our IMIL focuses on instance-level tasks with confounding issues caused by bag contextual prior.

\section{Interventional Multi-Instance Learning}

\subsection{Revisit MIL through Causal Inference}
Denote $\{X_i,Y_i\}_{i=1}^{I}$ as a set of coarsely-labeled bags. 
The bag $X_i=\{x_{ij}\}_{j=1}^{N_i}$ contains $N_i$ unlabeled instances --- each instance-level label $i.e.$, $y_{ij}\in\{0,1\}$ for $x_{ij}$, is unavailable. 
For each $X_i$, its bag-level label $Y_i$ is derived under the standard multiple instance assumption~\cite{SMI}, $i.e.$, $Y_i=1$ when $\exists y_{ij}=1$ for $j=1,...,N_i$, otherwise, $Y_i=0$.

\begin{figure}[t]
     \centering
     \subfigure[Causal graph]{
     \includegraphics[height=2.5cm]{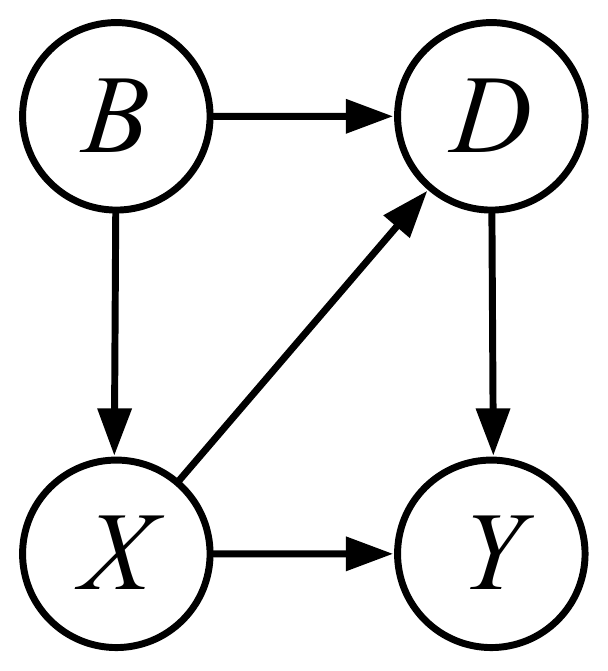}\label{fig: MIL SCM}
     }
     \quad\quad
     \subfigure[$P(Y|do(X))$]{
     \includegraphics[height=2.5cm]{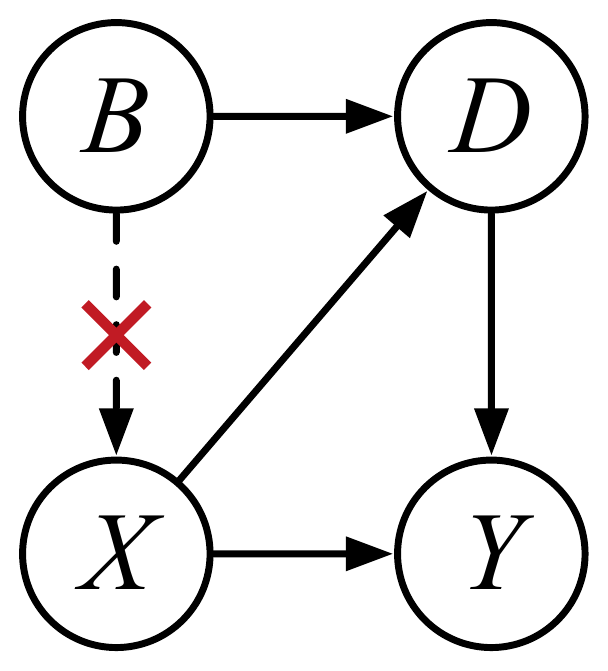}\label{fig: do(x)}
     }
    \caption{An illustration of the SCM model for MIL.
    }
    \label{SCM}
\end{figure}

Multi-instance learning aims at training a predictive model based on coarsely-labeled bags. 
As illustrated in Figure~\ref{fig: MIL SCM}, we can formulate the MIL framework as a causal graph (\textit{a.k.a}, a structural causal model or SCM)~\cite{pearl2016causal}, denoted as $\mathcal{G}=\{\mathcal{N}, \mathcal{E}\}$. 
The nodes $\mathcal{N}$ are a set of variables, and the edges $\mathcal{E}$ indicate causal relations in the system, which are shown below:
\begin{itemize}
    \item \textbf{$B\rightarrow X$:} We denote $X$ as the instance and $B$ as the bag contextual prior.
    This link reflects the fact that a bag contains multiple instances. 
    
    \item \textbf{$B \rightarrow D \leftarrow X$:} 
    We denote $D$ as the contextual information shared by the instances in the same bag ($a.k.a.$, instance-shared representation), derived based on the bag contextual prior $B$. 
    This contextual information is naturally encoded by MIL models as manifold bases~\cite{manifoldbase}, semantic topics~\cite{bau2017semantic}, typical patterns~\cite{zhang2018patterns}, $etc$. 

    \item \textbf{$X \rightarrow Y \leftarrow D$: }
    We denote $Y$ as the classification score determined by $X$ via a direct effect $X \rightarrow Y$ and an indirect effect $D\rightarrow Y$. 
    $X \rightarrow Y$ is obvious, which means the MIL model outputs $Y$ given $X$.
    On the other hand, $D \rightarrow Y$ indicates that the bag contextual prior affects the instance labels. 
    Note that $D \rightarrow Y$ always exists in MIL models. 
    Specifically, if $D \not \rightarrow Y$ in Figure~\ref{fig: MIL SCM}, the only path that transfers knowledge from $B$ to $Y$: $B \rightarrow X \rightarrow Y$ is blocked by conditioning on $X$ (d-separation~\cite{pearl2016causal}), then instance labels are no longer related to bags, which conflicts with the setting of MIL. 
\end{itemize}

Again, take WSI classification as an example:
1) \textbf{$B\rightarrow X$:} a WSI contains multiple patches, and the patches are of either different tissue types ($e.g.$ mitosis, cellular, neuronal and gland) or diagnoses ($e.g.$ cancer/non-cancer)~\cite{chan2019histosegnet}. 
2) \textbf{$B \rightarrow D \leftarrow X$:} The patches in a bag share some underlying information or features, $e.g.$, the global low-level features of the bag like colors and textures. 
Accordingly, $D$ corresponds to such instance-shared information. 
3) \textbf{$X \rightarrow Y \leftarrow D$:} A MIL model classifies the patches based on both instance-specific and instance-shared representations. 
Besides WSI classification, other MIL problems like temporal action localization~\cite{narayan20193c} (videos as bags and frames as instances) and weakly-supervised semantic segmentation~\cite{xu2019camel} (images as bags and objects as instances) can also be interpreted by the SCM in Figure~\ref{fig: MIL SCM}.

In our SCM graph, $B$ confounds $X$ and $Y$ via the backdoor path $ X\leftarrow B \rightarrow D\rightarrow Y$~\cite{backdoor}, $i.e.$, predicting all instances in a bag to be the same even if some instances are irrelevant to the prediction. 
On the other hand, $X \rightarrow D\rightarrow Y$ is a mediation path~\cite{pearl2013direct}, which is the key mechanism of MIL. 
Accordingly, the instance-shared information $D$ works as a mediator, which encodes the dependencies of instances. 
Take the objects in an indoor scene as an example.
An ``indoor'' bag tends to contain instances of ``TV'' rather than ``wild animal'', where $D$ contains the ``indoor'' semantics, which could narrow the search space for instance prediction $Y$ by filtering out those that belong to the ``outdoor'' scene. 
As detailed in the following sections, the main difference between our IMIL and the existing methods lies in the operations of confounder and mediator.

\subsection{Proposed Learning Method}
An ideal MIL model should capture the true causality between $X$ and $Y$. 
However, from the SCM in Figure~\ref{fig: MIL SCM}, the conventional correlation of $P(Y|X)$ fails to do so, because the likelihood of $Y$ given $X$ is not only due to $X$ \textit{per se}, but also the spurious correlation caused by the confounder $B$. 
Therefore, to pursue the true causality between $X$ and $Y$, we seek to use the causal intervention $P(Y|do(X))$ instead of the likelihood $P(Y|X)$ for MIL objective.
Here, the $do(\cdot)$ operation is defined as forcibly assigning a specific value to a variable, corresponding to applying random controlled trials~\cite{pearl2018book}.
Accordingly, we implement our IMIL framework by the following Expectation-Maximization (EM) algorithm, achieving deconfounded training and discriminative instance selection.
In the following content, we denote variable as a capital letter and denote value as  lowercase.

\subsubsection{M-step: Deconfounded training}
In M-step, the model is optimized under the physical intervention, which aims to ``cut-off'' the undesired confounding effect, as shown in Figure~\ref{fig: do(x)}. 
Instead of enumerating all possible instances in each bag, which is impossible in practice, we adopt strong data augmentations to mimic the random controlled trial. 
In particular, the bag context prior could be instantiated as structural patterns, geometric arrangements, color distributions, $etc$.
Applying data augmentations by  spatial and appearance transformations can enhance the diversity of the instances in the same bag. 
Accordingly, the augmented instances imitate the instances coming from different bags with various contextual priors, which achieves the $do(\cdot)$ operation exactly. 
In this work, we take the set of data augmentations used in MoCo v2~\cite{mocov2} as our default setting,\footnote{We call this setting ``strong'' augmentation in this work. The corresponding ``weak'' augmentation used in the following step just includes resizing, cropping, and flipping.} which includes resizing, cropping, horizontally flipping, color jittering, random grayscale conversion. Besides, we also consider random rotation, which has been widely used in image recognition tasks~\cite{rotation_related_wan2013regularization}.
As demonstrated in the following experiments, such deconfounded training brings significant improvements to all compared methods, which is a practical, generic, and implementation-friendly solution. 
It should be noted that leveraging the domain knowledge specialized by tasks may help design more effective data augmentation methods, which is left as our future work.

\subsubsection{E-step: Discriminative instance selection via total effect}

After applying the deconfounded training above in the M-step, we further select discriminative instances in the E-step, suppressing the confounding bias imposed by those non-discriminative instances in the next iteration. 
In principle, we introduce the Total Effect (TE) defined below as our criterion for instance selection:
\begin{eqnarray}\label{TE}
\begin{aligned}
TE(Y)&=\mathbb{E}[Y|do(X=x)]-\mathbb{E}[Y|do(X=x_0)]\\
&=\underbrace{P(1|do(X=x))-P(1|do(X=x_0))}_{\text{For binary MIL classification}},
\end{aligned}
\end{eqnarray}
which measures the expected effect on the prediction $Y$ as the instance $X$ changes from $x_0$ to $x$.
Here, $x_0$ is the reference instance, whose effect $\mathbb{E}[Y|do(X=x_0)]$ defines the boundary separating discriminative instances from non-discriminative ones.
Accordingly, we select the instances whose effects are larger than that of $x_0$. 
Obviously, the estimation of the TE consists of approximating the causal intervention $P(Y|do(X))$ and setting the reference effect $\mathbb{E}[Y|do(X=x_0)]$. 
In this work, we apply the backdoor adjustment~\cite{pearl2016causal} to achieve the causal intervention $P(Y|do(X))$ and propose a progressive curriculum to set the reference effect adaptively. 

At the $t$-th E-step, given a bag $\{x_{ij}\}_{j=1}^{N_i}$, we can calculate the score of each $x_{ij}$ by current model, denoted as $s^{(t)}(x_{ij})$. 
For MIL classification tasks, the score is often derived by a sigmoid or softmax operation.
Based on the scores, we approximate $\mathbb{E}[Y | do(X=x)]$ by an energy-based model~\cite{tang2020long}:
\begin{eqnarray}\label{eq:energy}
\begin{aligned}
\mathbb{E}[Y | do(X=x)] \propto \frac{S(x_{ij})}{\frac{1}{N_i}\sum_j S(x_{ij})} = E(x_{ij}).
\end{aligned}
\end{eqnarray}
This model is different from conventional softmax directly derived by the scores. 
Firstly, the $S(x_{ij})$ represents the exponential moving average scores derived by the Temporal Ensembling method~\cite{laine2016temporal}, which estimates the unnormalized effect of $do(X=x_{ij})$. 
It is calculated as follows:
\begin{equation}
    S(x_{ij}) \leftarrow m S(x_{ij})+(1-m) s^{(t)}(x_{ij}).
\end{equation}
This mechanism can be explained as applying momentary interval sampling multiple times, where $S(x_{ij})$ is the ensemble of scores and $m$ is a momentum coefficient. 
Applying $S(x_{ij})$ rather than $s^{(t)}(x_{ij})$ helps to enhance the robustness of instance selection~\cite{laine2016temporal}. 
Secondly, the denominator of $E(x_{ij})$ is the average score of all instances in a bag. 
It works as a propensity score~\cite{austin2011introduction}, balancing the observational bias of instances, as shown in the scatter plots of Figure~\ref{workflow}. 
Note that our implement is in the form of inverse probability weighting~\cite{pearl2016causal} since the confounder $B$ is unobserved.

For the reference effect, we initialize $\mathbb{E}[Y|do(X=x_0)]$ as $0$ and increase it gradually with the increase of iterations. 
In particular, we calculate $E(x_{ij})$ for each instance and remain the discriminative instances that correspond to the top $\lfloor R^{(t)}K\rfloor$ largest values in $\{E(x_{ij})\}$, where $R^{(t)} = (1-\tau t)$, $\tau$ is the decay ratio, and $K=\sum_{i=1}^{I}N_i$ is the number of instances. 
Accordingly, the reference effect at the $t$-th E-step, $i.e$,  $\mathbb{E}[Y|do(X=x_0^{(t)})]$, is
\begin{eqnarray}\label{eq:ref_effect}
\begin{aligned}
  \max\{\epsilon:  \sideset{}{_{x_{ij}}}\sum \mathbb{I}(E(x_{ij}) \geq \epsilon)=\lfloor R^{(t)}K\rfloor\},
\end{aligned}
\end{eqnarray}
where $\mathbb{I}(\cdot)$ is an indicator function generating $1$ if the statement is true.
The $R^{(t)}$ stops updating when the average reweighted score of the $\tau K$ smallest selected instances exceeds a predefined  threshold $T$, $i.e.$,
\begin{eqnarray}
\sideset{}{_{\mathcal{X}\subset \mathcal{X}_t}}\min  \Big\{ \frac{1}{\tau K} \sideset{}{_{x \in \mathcal{X}}}\sum E(x) : |\mathcal{X}| = \tau K \Big\} \geq T.
\end{eqnarray}
where $\mathcal{X}_{t}$ is the set of selected instances.
It is noted that the higher the score is, the more likely newly filtered instances are to be discriminative.  
Such a procedure can effectively prevent MIL models from over-fitting~\cite{wei2020combating}. 
Plugging Eq.~\eqref{eq:energy} and Eq.~\eqref{eq:ref_effect} into Eq.~\eqref{TE}, we approximate the TE for each instance $x$ at the $t$-th E-step as $TE(Y)=E(x)-\mathbb{E}[Y|do(X=x_0^{(t)})]$.

\subsection{Justifications}

\subsubsection{Implementations of Causal Intervention}
\begin{figure}[t]
    \centering
    \includegraphics[height=1.9cm]{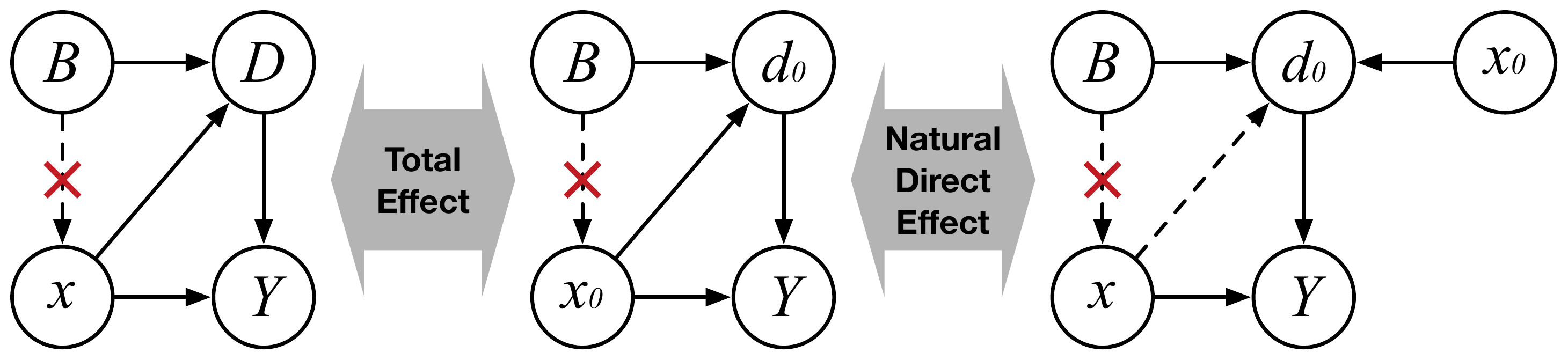}
    \caption{An illustration of the Total Effect and the Natural Direct Effect. 
    Here, $d_0$ is  the instance-shared information corresponding to the reference instance $x_0$.}
    \label{fig: TE}
\end{figure}

While causal intervention is agnostic to methods, datasets, and backbones in theory, their implementations are often task-specific in practice. 
Focusing on the EM framework of MIL, we implement the causal intervention at the M-step and that at the E-step by the physical intervention and the backdoor adjustment method, respectively. 
The physical intervention increases the diversity of data, which is beneficial for the M-step to avoid the over-fitting problem. 
However, it removes the confounding bias by introducing new randomness~\cite{sohn2020fixmatch} rather than selecting instances. 
On the contrary, the backdoor adjustment method approximates TE to select discriminative instances. 
As shown in Figure~\ref{fig: TE}, approximating TE helps to remove the undesired confounder bias ($i.e.$, cut off the backdoor path $ X\leftarrow B \rightarrow D\rightarrow Y$) and keep the mediator effect ($i.e$, reserve the mediator path $X \rightarrow D\rightarrow Y$), which is more suitable for the E-step.

\subsubsection{Multi-class Multi-label MIL Problems}
Our IMIL can be easily applied to multi-class multi-label MIL problems, as long as we take each class as a classical (binary) MIL problem. 
In such situations, the concept of ``discriminative'' varies from class, and the non-discriminative instance in one class may provide reliable supervision as a negative instance for another class. 
Therefore, we may select instances in a relatively conservative manner.
Specifically, if one instance is not selected for one class, we only reduce the significance of its loss in this class while maintaining its significance for the other classes.
This procedure can be understood as a more fine-grained soft version of our selection method.

\begin{table}[t]
\centering
\resizebox{0.88\columnwidth}{!}{
\begin{tabular}{ccccccc}
MIL & 
Simple & 
Patch & 
RCE & 
Top-\textit{k} & 
Semi & 
IMIL
\\ \midrule
\textit{do(D)} & 
- &
\CheckmarkBold &
\CheckmarkBold &
\CheckmarkBold &
- &
- \\
\textit{do(X)} & 
- &
- &
- &
- &
\CheckmarkBold &
\CheckmarkBold \\
Effect &
- & 
NDE &
NDE &
NDE &
TE &
TE       
\\ 
\end{tabular}
}
\caption{The differences among various MIL methods.}
\label{table1:methods}
\end{table}

\subsubsection{Connections with Existing MIL Methods}

It should be noted that our IMIL framework provides a new way to analyze existing state-of-the-art MIL methods. 
Specifically, Table~\ref{table1:methods} makes a comparison for various MIL methods from the viewpoint of causal intervention.
Essentially, most existing MIL methods can be categorized into three classes according to their instance selection strategies. 
The method in the first class is the SimpleMIL~\cite{SimpleMIL}, which simply uses all instances without causal intervention; 
The methods in the second class select instances by calculating the Natural Direct Effect (NDE) shown in Figure~\ref{fig: TE}. 
In particular, $NDE(Y) = \mathbb{E}[Y_{d_0}|do(X=x)]-\mathbb{E}[Y_{d_0}|do(X=x_0)]$, where $Y_{d_0}$ is the counterfactual output achieved under the condition of $do(X=x_0)$. 
Because the NDE completely removes the entire effect of $D$, which may lose some information beneficial for the learning problem, the MIL methods in this class often require some external information as compensations.
The representative methods include Top-\textit{k}MIL~\cite{chikontwe2020topk}, RCEMIL~\cite{RCE} and PatchCNN~\cite{hou2016cvpr}.
These methods select instances by comparing the scores of instances from the same bag, which is actually an intervention on $D$ that forces the mediator-specific effect to be the same. 
The $x_0$ is decided by the heuristic setting for Top-\textit{k}MIL, the statistical information for RCEMIL, and the bag-specific threshold for PatchCNN. 
The methods in the third class select instances by approximating the Total effect. 
Among them, the SemiMIL in~\cite{wang2019weakly} selects instances in a partially deconfounded manner, where the instances with extra annotations are adjusted by assigning a larger weight, while the rest is still confounded. 
Our proposed IMIL also uses TE for instance selection.
It uses the backdoor adjustment to separate the confounding effect from the mediator effect. 
This method can be taken as a plug-in component to the existing MIL methods.
Additionally, our IMIL is free of external information because TE retains the mediator effect of \textit{D},  which we believe is informative, and hence external information is unnecessary in our method.

\begin{table}[t]
\centering
\setlength{\tabcolsep}{4pt}
\resizebox{1\columnwidth}{!}{
\begin{tabular}{cccccc}
\multirow{3}{*}{Dataset} 
& \multirow{3}{*}{Class} 
& \multirow{2}{*}{Instance}
& \multicolumn{2}{c}{Train} 
& {Test} \\ 
\cmidrule(lr){4-5}  \cmidrule(lr){6-6} 
&  
& \multirow{2}{*}{Size}
& \# of
& \# of
& \# of\\
&
&
& bags
& instances
& instances\\
\midrule
\multirow{2}{*}{DigestPath} 
& Malignant 
& \multirow{2}{*}{512}
& 250 
& 10,133 
& N/A\\
& Normal 
&
& 410 
& 33,110 
& N/A \\ \midrule
\multirow{2}{*}{Camelyon} 
& Metastases 
& \multirow{2}{*}{256}
& 40 
& 64,430 
& 60,545 \\
& Normal 
&
& 40 
& 617,056 
& 60,545 \\
\end{tabular}}
\caption{Summary of the WSI datasets. 
N/A: not available.
}
\label{tab:dataset}
\end{table}

\begin{table*}[t]
\centering
\resizebox{1\textwidth}{!}{\begin{tabular}{crlcccccccccc} 
\multicolumn{3}{c}{\multirow{2}{*}{\textbf{Method}}} & \multicolumn{5}{c}{DigestPath} & \multicolumn{5}{c}{Camelyon} \\  \cmidrule(lr){4-8}   \cmidrule(lr){9-13} 
\multicolumn{3}{r}{} & AUC & ACC & F1 & REC & PRE & AUC & ACC & F1 & REC & PRE\\
\hline   
\multirow{11}{*}{\rotatebox{90}{\textbf{Using extra labels}}} 
&Oracle &  & 92.36 & 88.51 & 74.80 & 71.53 &80.63& 
81.35	&62.05	&39.48	&24.75	&97.46
\\  
& & + CI & 
\textbf{95.52}{\scriptsize\textcolor{ForestGreen}{+3.16}} & 
\textbf{90.66}{\scriptsize\textcolor{ForestGreen}{+2.15}} & \textbf{79.83}{\scriptsize\textcolor{ForestGreen}{+5.03}} & 76.85{\scriptsize\textcolor{ForestGreen}{+5.32}} & \textbf{84.99}{\scriptsize\textcolor{ForestGreen}{+4.36}} &
\textbf{86.23}{\scriptsize\textcolor{ForestGreen}{+4.88}}	&62.35{\scriptsize\textcolor{ForestGreen}{+0.30}}	&39.79{\scriptsize\textcolor{ForestGreen}{+0.31}}
&24.88{\scriptsize\textcolor{ForestGreen}{+0.13}}
&\textbf{99.29}{\scriptsize\textcolor{ForestGreen}{+1.83}}
\\ 
\cline{2-13}  
&RCEMIL &  & 87.50 & 82.75 & 63.90 & 65.21 & 63.83 
&79.52	&70.73	&62.95	&49.73	&85.75\\   
&& + CI & 
91.97{\scriptsize\textcolor{ForestGreen}{+4.47}} & 87.46{\scriptsize\textcolor{ForestGreen}{+4.71}} & 73.63{\scriptsize\textcolor{ForestGreen}{+9.73}} & 73.12{\scriptsize\textcolor{ForestGreen}{+7.91}} & 74.68{\scriptsize\textcolor{ForestGreen}{+10.85}} 
&85.48{\scriptsize\textcolor{ForestGreen}{+5.96}}	&76.09{\scriptsize\textcolor{ForestGreen}{+5.36}}	&69.55{\scriptsize\textcolor{ForestGreen}{+6.60}}	&54.60{\scriptsize\textcolor{ForestGreen}{+4.87}}	&95.77{\scriptsize\textcolor{ForestGreen}{+10.02}}\\ 
&& + CI$*$ & 
91.65{\scriptsize\textcolor{RedOrange}{-0.32}} & 
85.58{\scriptsize\textcolor{RedOrange}{-1.88}} & 
71.13{\scriptsize\textcolor{RedOrange}{-2.50}} & 
69.14{\scriptsize\textcolor{RedOrange}{-3.98}} & 
74.90{\scriptsize\textcolor{ForestGreen}{+0.22}} &
84.69{\scriptsize\textcolor{RedOrange}{-0.79}} &
76.98{\scriptsize\textcolor{ForestGreen}{+0.89}}&
71.29{\scriptsize\textcolor{ForestGreen}{+1.74}}&
57.15{\scriptsize\textcolor{ForestGreen}{+2.55}}&
94.70{\scriptsize\textcolor{RedOrange}{-1.07}}
\\ 
\cline{2-13}   
&PatchCNN &  & 91.09 & 83.46 & 70.26 & 81.85 & 62.86 
&78.61	&73.24	&67.61	&55.84	&85.65\\   
&& + CI & 
94.53{\scriptsize\textcolor{ForestGreen}{+3.44}} & 88.28{\scriptsize\textcolor{ForestGreen}{+4.82}} & 77.92{\scriptsize\textcolor{ForestGreen}{+7.66}} & 85.69{\scriptsize\textcolor{ForestGreen}{+3.84}} & 72.05{\scriptsize\textcolor{ForestGreen}{+9.19}} &
85.42{\scriptsize\textcolor{ForestGreen}{+6.81}}	&\textbf{80.78}{\scriptsize\textcolor{ForestGreen}{+7.54}}	&\textbf{77.11}{\scriptsize\textcolor{ForestGreen}{+9.50}}	&64.75{\scriptsize\textcolor{ForestGreen}{+8.91}}	&95.30{\scriptsize\textcolor{ForestGreen}{+9.65}}\\   
& & + CI$*$ & 
91.84{\scriptsize\textcolor{RedOrange}{-2.69}} & 
82.13{\scriptsize\textcolor{RedOrange}{-6.15}} & 
69.92{\scriptsize\textcolor{RedOrange}{-8.00}} & 
\textbf{88.29}{\scriptsize\textcolor{ForestGreen}{+2.60}} & 
57.57{\scriptsize\textcolor{RedOrange}{-14.48}} &
71.06{\scriptsize\textcolor{RedOrange}{-14.36}}&
67.81{\scriptsize\textcolor{RedOrange}{-12.97}}&
65.63{\scriptsize\textcolor{RedOrange}{-11.48}}&
61.47{\scriptsize\textcolor{RedOrange}{-3.28}} &
70.40{\scriptsize\textcolor{RedOrange}{-24.90}}

\\ 
\cline{2-13}   
&SemiMIL &  & 91.94 & 87.68 & 75.43 & 78.28 & 74.02
&72.2  &67.26   &66.63   &65.38   &67.94\\  
& & + CI & 
94.40{\scriptsize\textcolor{ForestGreen}{+2.46}} & 
90.10{\scriptsize\textcolor{ForestGreen}{+2.42}} & 
79.90{\scriptsize\textcolor{ForestGreen}{+4.47}} & 
80.83{\scriptsize\textcolor{ForestGreen}{+2.55}} & 80.07{\scriptsize\textcolor{ForestGreen}{+6.05}} 
&77.49{\scriptsize\textcolor{ForestGreen}{+5.29}}
&72.99{\scriptsize\textcolor{ForestGreen}{+5.73}}
&71.35{\scriptsize\textcolor{ForestGreen}{+4.72}}
&\textbf{67.28}{\scriptsize\textcolor{ForestGreen}{+1.90}} &75.95{\scriptsize\textcolor{ForestGreen}{+8.01}}\\  
& & + CI$*$ & 
92.81{\scriptsize\textcolor{RedOrange}{-1.59}} & 
85.88{\scriptsize\textcolor{RedOrange}{-4.22}} & 
73.98{\scriptsize\textcolor{RedOrange}{-5.92}} & 
85.28{\scriptsize\textcolor{ForestGreen}{+4.45}} & 65.76{\scriptsize\textcolor{RedOrange}{-14.31}} &
73.24 {\scriptsize\textcolor{RedOrange}{-4.25}}&
69.89{\scriptsize\textcolor{RedOrange}{-3.10}}&
67.77{\scriptsize\textcolor{RedOrange}{-3.58}}&
63.30{\scriptsize\textcolor{RedOrange}{-3.98}}&
72.92{\scriptsize\textcolor{RedOrange}{-3.03}}\\ 
\hline 
\multirow{6}{*}{\rotatebox{90}{\textbf{No extra labels}}}
&SimpleMIL &  & 88.46 & 77.20 & 64.70 & 87.95 & 51.60 
&72.44	&67.00	&66.98	&\textbf{66.95}	&67.01\\   
&& + CI & 
91.05{\scriptsize\textcolor{ForestGreen}{+2.59}} & 
79.70{\scriptsize\textcolor{ForestGreen}{+2.50}} & 
67.30{\scriptsize\textcolor{ForestGreen}{+2.60}} & 
\textbf{89.09}{\scriptsize\textcolor{ForestGreen}{+1.14}} & 
54.30{\scriptsize\textcolor{ForestGreen}{+2.70}} &
73.81{\scriptsize\textcolor{ForestGreen}{+1.37}}
&68.05{\scriptsize\textcolor{ForestGreen}{+1.05}}	&67.42{\scriptsize\textcolor{ForestGreen}{+0.44}}	&66.12{\scriptsize\textcolor{RedOrange}{-0.83}}	&68.77{\scriptsize\textcolor{ForestGreen}{+1.76}}\\ 
\cline{2-13}  
&Top-$k$MIL &  & 88.57 & 80.97 & 66.88 & 80.87 & 58.30 
&66.89	&57.10	&26.27	&15.28	&\textbf{93.42}\\  
&& + CI & 
\textbf{93.01}{\scriptsize\textcolor{ForestGreen}{+4.44}} & 85.75{\scriptsize\textcolor{ForestGreen}{+4.78}} & \textbf{73.93}{\scriptsize\textcolor{ForestGreen}{+7.05}} & 85.32{\scriptsize\textcolor{ForestGreen}{+4.45}} & 65.55{\scriptsize\textcolor{ForestGreen}{+7.25}} &
69.22{\scriptsize\textcolor{ForestGreen}{+2.33}}	&63.99{\scriptsize\textcolor{ForestGreen}{+6.89}}	&48.17{\scriptsize\textcolor{ForestGreen}{+21.9}}	&33.47{\scriptsize\textcolor{ForestGreen}{+18.19}}	&85.93{\scriptsize\textcolor{RedOrange}{-7.49}}\\ 
\cline{2-13}   
&IMIL (Ours) &  & 88.16 & 81.09 & 65.81 & 77.83 & 57.65 
&74.19	&69.85	&67.33	&62.13	&73.47 \\  
&& + CI & 
92.06{\scriptsize\textcolor{ForestGreen}{+3.90}} & 
\textbf{86.84}{\scriptsize\textcolor{ForestGreen}{+5.75}} & 73.74{\scriptsize\textcolor{ForestGreen}{+7.93}} & 78.77{\scriptsize\textcolor{ForestGreen}{+0.94}} & \textbf{69.6}{\scriptsize\textcolor{ForestGreen}{+11.95}} &
\textbf{80.18}{\scriptsize\textcolor{ForestGreen}{+5.99}} 	&\textbf{75.34}{\scriptsize\textcolor{ForestGreen}{+5.39}} 	&\textbf{71.35}{\scriptsize\textcolor{ForestGreen}{+4.02}} 	&61.43{\scriptsize\textcolor{RedOrange}{-0.70}} 	&85.10{\scriptsize\textcolor{ForestGreen}{+11.63}} \\ 
\end{tabular}} 
\caption{Numerical results (\%) on dataset of DigestPath and Camelyon16. 
`+CI' means that causal intervention is applied in the M-step. `*' means the external information is noised.
}
\label{tab:main results}
\end{table*}
\section{Experiments}
\subsection{Pathological Image Analysis}
To demonstrate the effectiveness of our IMIL method, we apply it to the WSI classification problem and compare it with state-of-the-art MIL methods. 
In particular, we consider the classification of colonoscopy tissues (malignant v.s. normal) and the classification of lymph node sections (metastases v.s. normal).
The datasets used in our experiments are DigsetPath~\cite{digestpath}\footnote{\url{https://digestpath2019.grand-challenge.org/}} and Camelyon16~\cite{CAMELYON16}\footnote{\url{https://camelyon16.grand-challenge.org/}}, both of which have bag-level and instance-level labels for each image and its patches, respectively. 
Specifically, we apply Otsu's method~\cite{otsu1979threshold} to remove the background of the images and extract non-overlapped patches from the foreground regions. 
The statistics of these two datasets are summarized in Table~\ref{tab:dataset}. 

The competitors of our method include an \textbf{Oracle} model with full supervision at the instance-level and the state-of-the-art MIL methods like the \textbf{SimpleMIL} in~\cite{SimpleMIL}, the \textbf{PatchCNN}) in~\cite{hou2016cvpr}, the \textbf{RCEMIL} trained by RCE loss~\cite{RCE}, the \textbf{SemiMIL} trained with both bag-level and partial instance-level labels, and the \textbf{Top-$k$MIL} trained by Top-$k$ selection~\cite{chikontwe2020topk}.  
For PatchCNN, SemiMIL and RCEMIL, some external information should be provided: the specific threshold for each bag, partial instance-level labels, and two statistical values for the re-weighting scheme. 
For other methods, including ours, only the bag-level labels are used for training.

For a fair comparison, all the methods use  ResNet-18~\cite{ResNet} as their backbone models. 
Adam optimizer is used with an initial learning rate of 0.001, and the batch size is set to 64. 
We run 50 epochs in total and decay the learning rate with the cosine decay schedule~\cite{loshchilov2016sgdr}. 
For our method, the hyperparameters are $m=0.5$, $\tau =0.05$ and $T= 0.95$ by default.
We evaluate the instance-level performance of each method based on 5-fold cross-validation, and the measurements include Area Under Curve (AUC), accuracy (ACC), F1-score, recall (REC) and precision (PRE). 
The numerical comparisons for various methods on the two datasets are shown in Table~\ref{tab:main results}. 
We can find that the causal intervention consistently improves the methods in all settings, indicating that causal intervention is agnostic to methods and datasets.

For the methods using external information, the finer granularity the supervision used, the better performance is achieved. 
However, their performance is also highly dependent on the quality of external labels, which decreases dramatically given noisy external labels (denoted by $*$). 
Note that the results of the oracle model should indicate the upper bound of obtainable performance, while its low recall on Camelyon16 is mainly due to the highly imbalanced ratio of the data. 
Among the methods without external information, our IMIL obtains superior performance. 
It consistently improves SimpleMIL on most measurements by non-trivial margins. 
Especially, IMIL remarkably promotes precision by $15.3\%$ and $16.33\%$ on DigestPath and Camenlyon16, respectively, indicating a substantial reduction over false positives. 
Although the method of Top-$k$MIL performs well on DigetPath, it receives extremely low recall ($15.28\%$ without causal intervention) on Camelyon16 because the setting of $k$ is sensitive to the change of bag size.
Additionally, the top-\textit{k} selection may degenerate into max-max selection criteria, which tends to have relatively low recall and high specificity~\cite{xu2019camel}. 
On the contrary, our method can adaptively select  discriminative instances instead of setting a fixed number.

\begin{figure}[t]
    \centering
    \includegraphics[width=0.8\columnwidth]{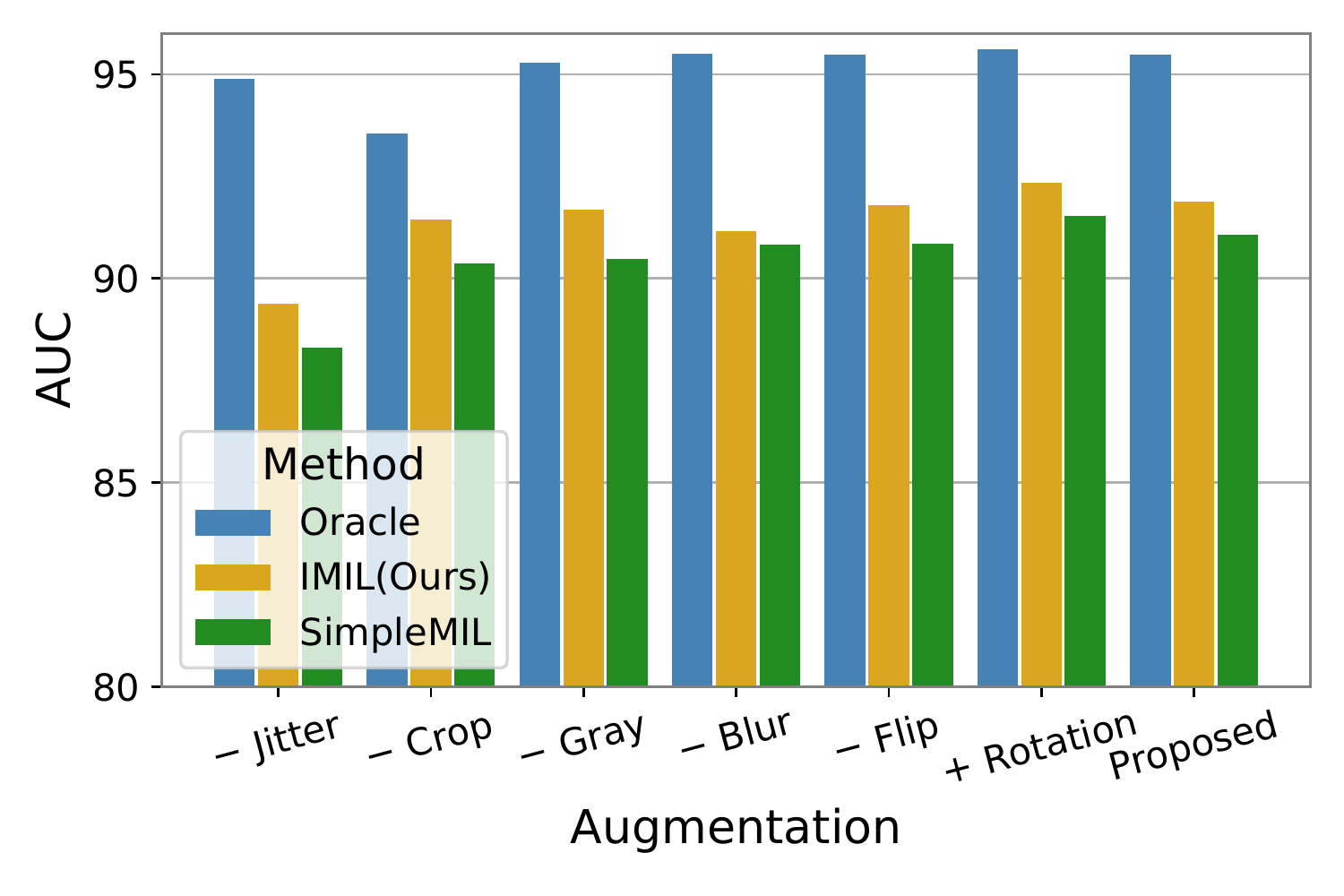}
    \caption{AUC under different composition of data augmentations, where $\mathbf{-/+}$ means one individual data augmentation is removed/added. 
    }
    \label{AUG}
\end{figure}

\begin{table}[t]
\centering
\resizebox{1\columnwidth}{!}{
\begin{tabular}{cl|ccccc}
Parameters &Settings & AUC & Acc & F1& REC & PRE \\ \midrule
&$m$ = 0 & 92.18 & 86.69 & 73.55 & 79.25 & 68.80 \\
&$m$ = 0.25 & 91.97 & 86.72 & 73.37 & 78.16 & 69.25 \\
Momentum &$m$ = 0.5 & 92.06 & 86.84 & 73.74 & 78.77 & 69.60 \\
$m$ &$m$ = 0.75 & \textbf{92.47} & \textbf{87.32} & \textbf{74.68} & \textbf{79.32} &\textbf{ 70.70 }\\
&$m$ = 0.9 & 91.60 & 85.56 & 72.88 & 79.75 & 68.07 \\ \midrule
 & $T $ = 0.9 & 91.67 & 86.69 & 72.66 & 76.34 & 69.55 \\
Threshold & $T$ = 0.95 & 92.06 & 86.84 & 73.74 & \textbf{78.77} & 69.60 \\
$T$ & $T $ = 1 & \textbf{92.23} & \textbf{87.35} & \textbf{74.34} & 77.94 & \textbf{71.46 }\\
& $T $ = 1.05  & 91.28 & 86.79 & 72.23 & 74.12 & 70.49 \\ \midrule
&$\tau$ = 0.025 & 91.99 & 86.90 & 73.88 & 78.02 & 70.46 \\
Step &$\tau$ = 0.05 & 92.06 & 86.84 & 73.74 & \textbf{78.77 }& 69.60 \\
$\tau$ &$\tau$ = 0.075 &  \textbf{92.46 }& \textbf{87.43} & \textbf{74.56} & 78.20 & \textbf{71.46}\\
\end{tabular}
}
\caption{The robustness of our IMIL to its hyperparameters.}
\label{hyperparameter}
\end{table}

\subsection{Further Analysis}
\subsubsection{Data Augmentations}
To quantitatively assess the contributions of different data augmentations, we remove/add individual data augmentation in the M-step on three methods ($i.e.$, Oracle, SimpleMIL, and our IMIL). 
For these methods, removing/adding a significant augmentation method is expected to harm/improve their performance. 
Figure~\ref{AUG} indicates that ``color jittering'' is the most important for IMIL and SimpleMIL --- the instances in the same bag share both labels and staining conditions, thus jittering could prevent their models from exploiting this spurious correlation.
For the oracle model, the ``color jittering'' is unimportant since the instances are fully-supervised, and the staining condition will cause less confounding bias. 
Instead, the models may over-fit the co-occurrence of tissues, thus ``random resizing and cropping'' is necessary for removing these biases. 
For other augmentations, ``grayscale conversion'' and ``Gaussian blurring'' achieve the causal intervention at the cost of hurting the characteristics (the HE staining and the resolution) of WSIs, while ``flipping'' and ``rotation'' bring limited improvements because WSIs have no dominant orientation.

\subsubsection{Robustness to Hyperparameters}
For those important hyperparameters of our method, we change one of them in a range and fix the others to their default values. 
The performance of our IMIL under different configurations is shown in Table~\ref{hyperparameter}. 
For the momentum used in Temporal Ensembling, the best performance is achieved when $m=0.75$. This result reflects that a relatively large momentum is beneficial for robust instance selection.
For the threshold used in selecting curriculum, we set $T$ around $1$ for the assumption that scores of positive and negative instances can be separated by average scores in most case, as shown in Figure~\ref{workflow}. 
Intuitively, higher threshold will keep less instances selected and result in lower recall ($T=1.05$). 
For the step used in selecting curriculum, a small $\tau$ results in a slow process of instance selection, while a large one may be too aggressive. Overall, our IMIL is robust to the hyperparameters.

\subsubsection{Qualitative Results}
\begin{figure}[t]
     \centering
     \subfigure[]{
     \includegraphics[height=2.3cm]{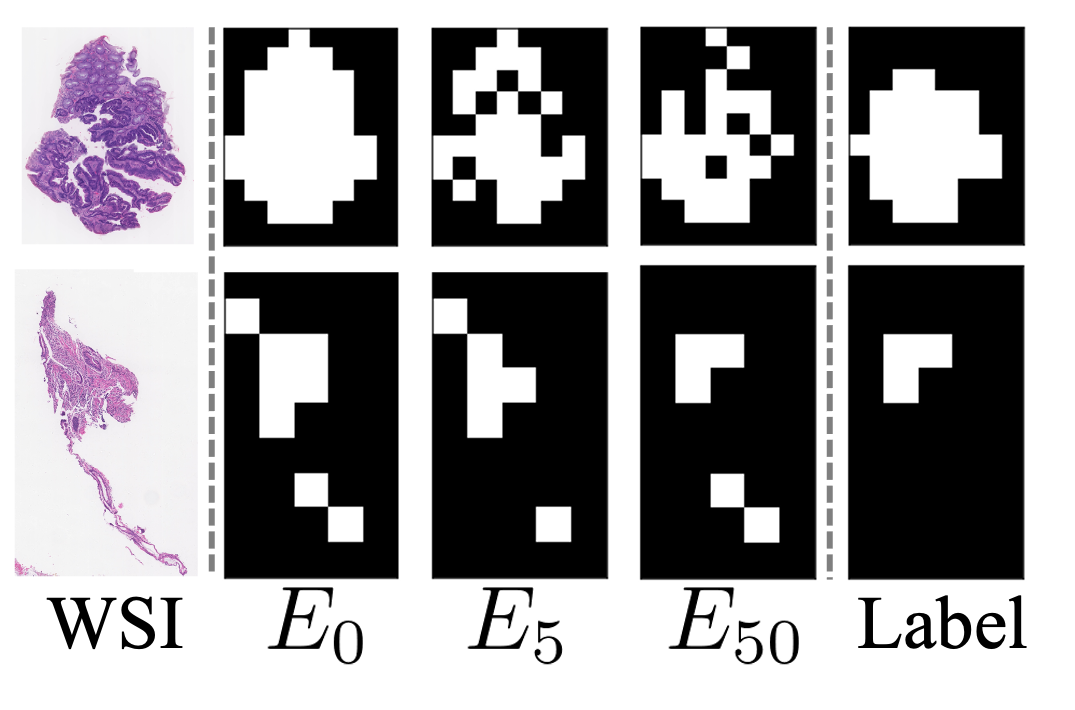}\label{fig: visa}
     }
     \subfigure[]{
     \includegraphics[height=2.25cm]{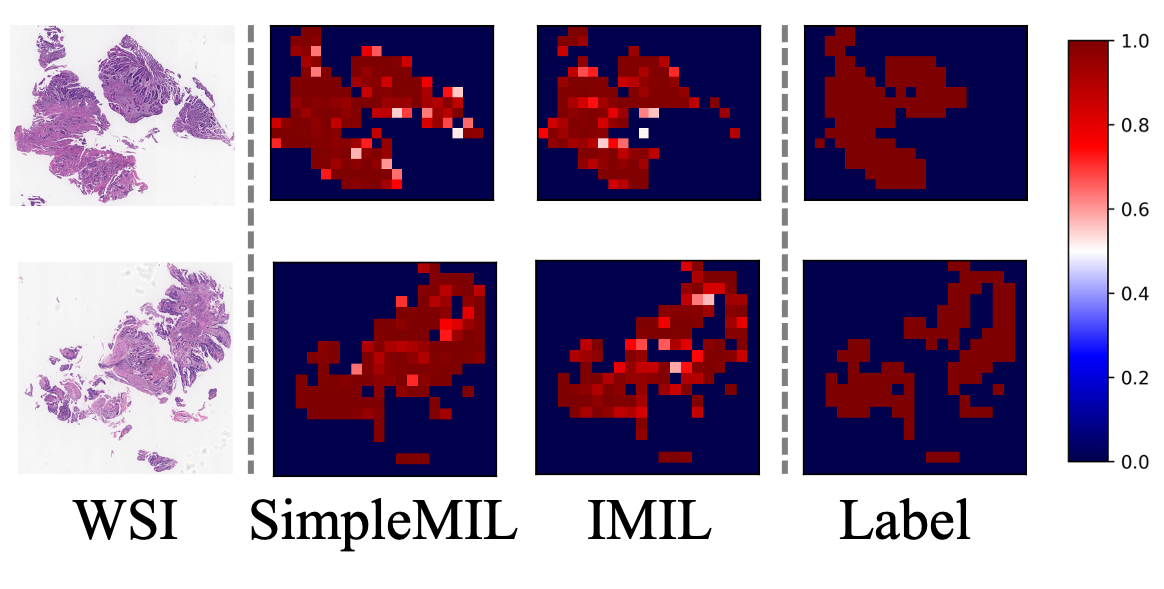}\label{fig: visb}
     }

        \caption{(a) The procedure of selecting curriculum: From left to right, WSI, selected instances at epoch 0/5/50, denoted by $E_0$, $E_5$ and $E_{50}$, and ground-truth of instance label. 
        (b) The results of instance classification: From left to right, WSI, result of SimpleMIL, result of IMIL, ground-truth of instance label.}
\end{figure}

We present qualitative results of  IMIL in two aspects: the selecting curriculum procedure and the patch classification on new WSIs. 
In Figure~\ref{fig: visa}, discriminative instances are gradually selected, validating our initial assertions that the average scores of bags naturally can serve as propensity scores.  Though the achieved supervision is not perfectly clean, it is worth noting that we do not need any external supervision. 
In Figure~\ref{fig: visb}, the heatmaps indicate the regions of high tumor probability. 
Notably, our IMIL can accurately distinguish tumors from normal tissues and considerably reduce false positives compared to SimpleMIL, suggesting the effectiveness of the proposed method.

\subsection{Bag-Level Classification on Pascal VOC}
To demonstrate the universality of our IMIL method, besides WSI classification problems, we further test it on the PASCAL VOC 07 dataset~\cite{everingham2015pascal}. 
This dataset contains 9963 natural images of 20 categories, which is challenging as the appearances of objects are diverse.
Following~\cite{deep_mil}, we take each image as a bag and adopt region proposal methods, $e.g$, Region Proposal Network (RPN)~\cite{ren2015faster}, to generate instances. 
Since the instance-level labels are unavailable in VOC, we only deploy the methods free of extra information and equip them with three MIL pooling operators, $i.e.$, max pooling, mean pooling and voting, for bag classification. 
For all the methods, we reports its mean average precision (mAP) on the \textit{test set} in Table~\ref{pascal}. 
The proposed causal intervention steadily improves all compared methods, where our IMIL achieves the best performance on bag-level prediction, demonstrating the proposed framework's stability and effectiveness. 

\begin{table}[t]
\centering
\resizebox{0.85\columnwidth}{!}{
\begin{tabular}{ccccc}
\multicolumn{2}{c}{\multirow{2}{*}{Method}} & \multicolumn{3}{c}{Aggregator} \\ \cline{3-5} 
\multicolumn{2}{c}{} & max & mean & voting \\ \hline
SimpleMIL & \multicolumn{1}{c}{} & 63.89 &	68.2 & 65.46 \\
 & \multicolumn{1}{c}{+CI} & 74.8{\scriptsize\textcolor{ForestGreen}{+10.91}}
 & 77.97 {\scriptsize\textcolor{ForestGreen}{+9.77}}& 75.46 {\scriptsize\textcolor{ForestGreen}{+10.0}}\\ \hline
Top-$k$MIL & \multicolumn{1}{c}{} & 64.77 & 67.97 & 65.13 \\
 & \multicolumn{1}{c}{+CI} & 73.48{\scriptsize\textcolor{ForestGreen}{+8.71}} & 77.13{\scriptsize\textcolor{ForestGreen}{+9.16}} & 74.74 {\scriptsize\textcolor{ForestGreen}{+9.61}}\\ \hline
IMIL & \multicolumn{1}{c}{} & 65 & 68.65 & 65.66 \\
 & \multicolumn{1}{c}{+CI} & \textbf{75.78}{\scriptsize\textcolor{ForestGreen}{+10.78}} & 
\textbf{ 78.34} {\scriptsize\textcolor{ForestGreen}{+9.69}}& \textbf{75.53}{\scriptsize\textcolor{ForestGreen}{+9.87}} \\ 
\end{tabular}
}
\caption{The mean average precision over Pascal VOC 07.}
\label{pascal}
\end{table}

\section{Conclusion}
We present a novel interventional multi-instance learning (IMIL) method to suppress the negative influence of bag contextual prior to instance-level prediction. 
In particular, we propose a causal graph for MIL and equip the EM-based MIL paradigm with causal intervention, combining the training process with data augmentations and adaptive instance selection.
Experimental results show that our IMIL achieves promising performance on various computer vision tasks.
In the future, we plan to design more task-specific data augmentation methods to improve the physical intervention strategy. 
Moreover, a systematic comparison  will be considered for various causal intervention methods in MIL tasks. 

\section{Acknowledgments}
Dr. Yi Xu was supported in part by NSFC 62171282, Shanghai Municipal Science and Technology Major Project (2021SHZDZX0102), 111 project BP0719010, and SJTU Science and Technology Innovation Special Fund ZH2018ZDA17. 
Dr. Hongteng Xu was supported in part by NSFC 62106271 and Intelligent Social Governance Platform, Major Innovation \& Planning Interdisciplinary Platform for the ``Double-First Class'' Initiative, RUC. 
He also thanks the support of Public Policy and Decision-making Research Lab of RUC and JD Explore Academy.
Authors thank Zengchao Xu for inspiring discussions.

\bibliography{reference}

\end{document}